\documentclass[letterpaper, 10 pt, conference]{ieeeconf}  

\IEEEoverridecommandlockouts                              

\usepackage{amsmath,amsfonts,amssymb}
\usepackage{mathrsfs}
\usepackage{algorithmic}
\usepackage{algorithm}
\usepackage{array}
\usepackage[caption=false,font=normalsize,labelfont=sf,textfont=sf]{subfig}
\usepackage{textcomp}
\usepackage{stfloats}
\usepackage{url}
\usepackage{verbatim}
\usepackage{graphicx}
\usepackage{cite}
\usepackage{bm}
\usepackage{float}
\usepackage{balance}
\usepackage{tikz}
\usepackage{ulem}
\title{\LARGE \bf
In-between Motion Generation Based Multi-Style Quadruped Robot Locomotion
}

\author{Yuanhao Chen$^{\dagger}$, 
Liu Zhao$^{\dagger}$,
Ji Ma,
Peng Lu$^{*}$
\thanks{$^\dagger$ Equal Contribution.}
\thanks{$^{*}$ Corresponding author: lupeng@hku.hk}
\thanks{The authors are with the Adaptive Robotic Controls Lab (Arclab), Department of Mechanical Engineering, The University of Hong Kong, Hong Kong SAR, China, jyhcan@connect.hku.hk, zhaol@connect.hku.hk, maji@connect.hku.hk} 
}

\begin{document}



\maketitle
\thispagestyle{empty}
\pagestyle{empty}



\begin{abstract}


Quadruped robots face persistent challenges in achieving versatile locomotion due to limitations in reference motion data diversity. To address these challenges, we introduce an in-between motion generation based multi-style quadruped robot locomotion framework. We propose a CVAE based motion generator, synthesizing multi-style dynamically feasible locomotion sequences between arbitrary start and end states. By embedding physical constraints and leveraging joint poses based phase manifold continuity, this component produces physically plausible motions spanning multiple gait modalities while ensuring kinematic compatibility with robotic morphologies. We train the imitation policy based on generated data, which validates the effectiveness of generated motion data in enhancing controller stability and improving velocity tracking performance.
The proposed framework demonstrates significant improvements in velocity tracking and deployment stability. We successfully deploy the framework on a real-world quadruped robot, and the experimental validation confirms the framework's capability to generate and execute complex motion profiles, including gallop, tripod, trotting and pacing.

\end{abstract}

\section{INTRODUCTION}


In recent years, the development of quadruped robots has attracted significant attention within the realm of robotics. Quadruped robot motion control, particularly for complex dynamic gaits, remains a significant challenge. While traditional reinforcement learning suffers from limited guidance provided by the reward design, imitation learning using real-dog motion capture data offers a promising path toward naturalistic movement \cite{schaal2003computational,kober2010imitation,bin2020learning}.

However, reliance on such data introduces critical limitations to imitation learning: acquisition requires specialized facilities, yielding datasets that are scarce, short, and velocity-incomplete \cite{hua2021learning}. This scarcity of motion capture data directly undermines policy performance, leading to poor velocity tracking and instability.

To address these issues, researchers have started to explore different approaches that can learn the style of a real dog's movements from motion capture data. \cite{li2024fld} encoded motion 
\begin{figure}[htbp] 
    \centering
    \subfloat[Gallop]{ 
    \begin{minipage}[b]{0.5\textwidth}
        \centering  
        \includegraphics[width=\linewidth]{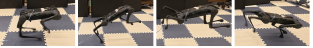} 
    \end{minipage}} 
    \\
       \subfloat[Tripod]{ 
    \begin{minipage}[b]{0.5\textwidth}
        \centering  
        \includegraphics[width=\linewidth]{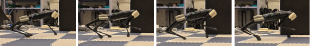} 
    \end{minipage}} 
    \caption{\textbf{Deployment result}. Gallop and Tripod motion that learned from the in-between motion generated motion.} 
    \label{Figure: Result of AMP in real world} 
\end{figure}
styles via latent representations, enabling bipedal robots to learn diverse movement patterns. \cite{peng2021amp,escontrela2022adversarial} utilized adversarial motion priors (AMP) for imitation learning, yet remains constrained by dataset scarcity. Although these methodologies represent fundamental advances in the treatment of these issues, the core challenge of the lack of data in motion capture remains unexplored.

To overcome the data scarcity challenge, we develop an in-between motion generation algorithm specifically designed for quadruped robotic configurations for imitation learning. We propose an in-between motion generation based multi-style quadruped robot locomotion framework, which utilizes observable data from quadruped simulation environments as both input and output, strictly adhering to joint constraints from robot physical capabilities. Furthermore, we redesign the loss functions, making them adapted to joint limits and constraints for implementation objectives. Moreover, a first-frame prediction is proposed for deployment. A generated motion based AMP is used to validate the effectiveness of the generated data. Results has been validated in the real-world deployment shown in Fig. \ref{Figure: Result of AMP in real world}.

\section{RELATED WORK}

\subsection{Motion In-between Generation}

In the early research, the In-between motion generation problem was often described as a motion planning problem, employing methods like motion graphs \cite{arikan2002interactive,beaudoin2008motion}, optimization \cite{chai2007constraint}, and constraint-limited A* search \cite{safonova2007construction}. Deep learning has now emerged as a prevalent tool in addressing the in-between motion generation problem.  \cite{harvey2020robust} built up their model using recurrent neural networks (RNNs) to predict the motion and position of the next frame with messages and constraints in the past.  \cite{tang2022real} used a conditional variational autoencoder (CVAE) network. This network is able to generate motions that match the character's movement speed, which avoids foot skating. Based on \cite{tang2022real}, \cite{song2022rsmt} combined the PAE network from \cite{starke2022deepphase} and \cite{tang2022real}, creating a real-time stylized motion transition model that can provide in-between motion with different styles. \cite{KIM2022108894} resolved the lack of semantic control and pose-specific controllability in traditional Motion In-Betweening (MIB) tasks. \cite{10.1145/3641519.3657414} solved the inflexibility of prior methods. \cite{Yun_2025_CVPR} integrated video diffusion models, ICAdapt fine-tuning, and motion-video mimicking to achieve in-between motion generation for arbitrary characters. 

\subsection{Imitation Learning Controller}

Lots of robot control policies that can be implemented on real robots have used imitation learning methods for training. For those that did not use the motion data directly, \cite{siekmann2021sim} implemented a parametric reward function for all common bipedal gaits, demonstrating a successful sim-to-real transfer. \cite{escontrela2022adversarial} proposed an adversarial motion priors (AMP) to replace the complex reward function in control policy training. \cite{shao2021learning,li2024fld} leveraged the phase vector as a guide for training. \cite{shao2021learning} proposed a network based on phase periodic reward function and \cite{li2024fld} offered a motion learning controller with fourier latent dynamics. \cite{10499824} provided an AI-CPG method, learning a combination of central pattern generators and generate humanoid locomotion.

For those who used the motion data directly instead of using a latent vector, \cite{ji2024exbody2} implemented a teacher-student policy learning policy with a CVAE network to expand the dataset for imitation learning. \cite{he2025asap} solved the sim-to-real dynamics gap by proposing the ASAP framework with a two-stage training to enable agile whole-body control.

\section{METHODOLOGY}

\begin{figure*}[htbp]
    \centering
	\centerline{\includegraphics[width=0.92\linewidth]{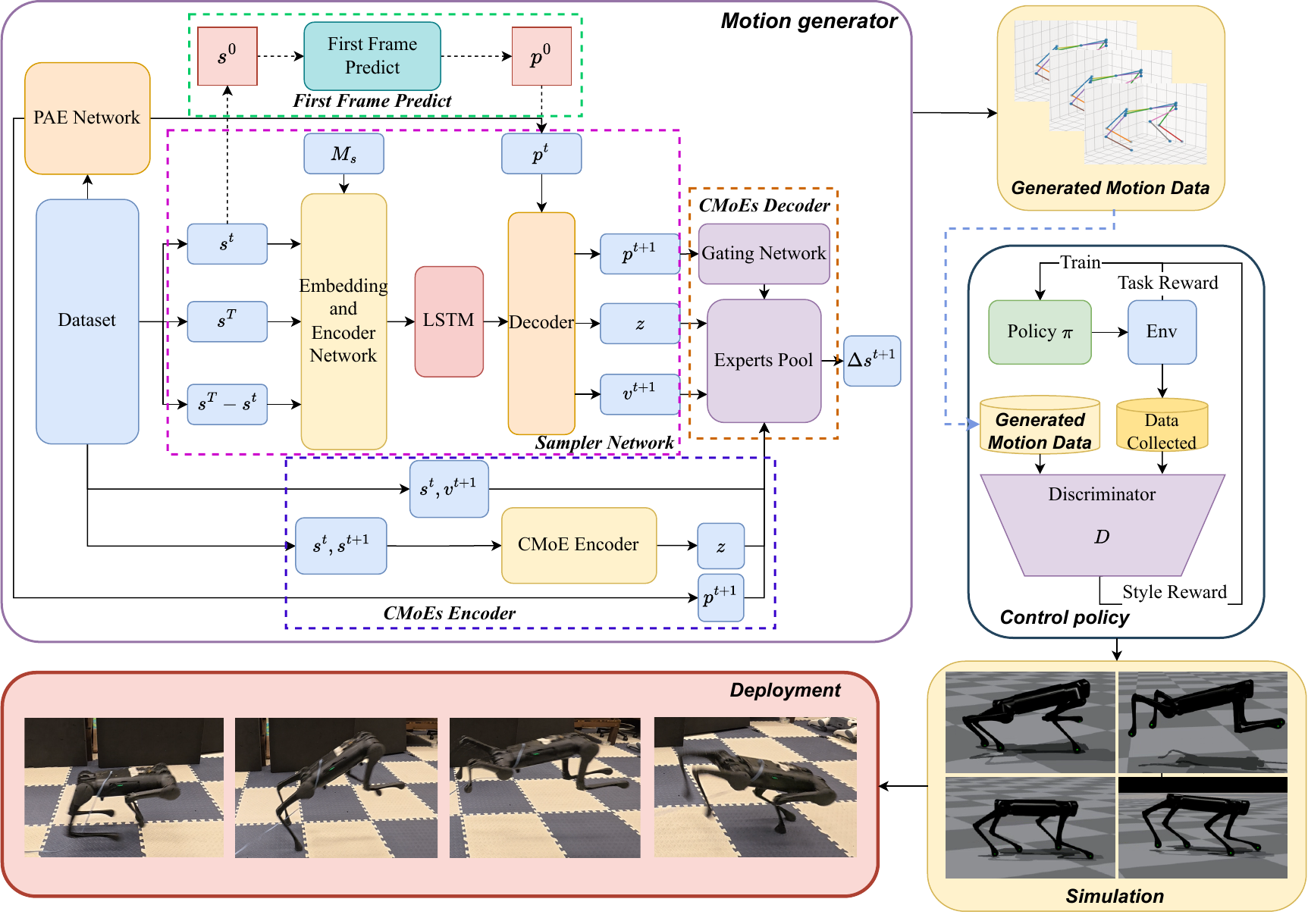}}
	\caption{\textbf{framework Overview}. This framework comprises two key components: a Motion Generator and a Control Policy. Motion trajectories generated by the Motion Generator serve as reference motions for Imitation Learning. Simulation results illustrate the policy’s learned behavior. Finally, Deployment demonstrates the GALLOP motion successfully transferred to the real-world hardware.}
	\label{Figure: outline}
\end{figure*}

Our framework implements a physically constrained motion generator that adheres to robotic mechanical limitations. By leveraging motion generation to diversify the original training dataset and establishing a unified controller-integrated motion synthesis architecture, we achieve enhanced velocity tracking performance with accuracy and adaptability in robotic locomotion control systems. The framework overview is shown in Fig. \ref{Figure: outline}.

\subsection{Quadruped Adaptive In-between Motion Generator}

\subsubsection{Data Formatting}



In this approach, to ensure that the generated motion adheres to the physical constraints on a real robot, we implement a differentiated data formatting strategy. Considering robot root stability, physical constraints, and quadruped robot dynamics, the robot's root joint is processed separately from its other joints during motion generation input construction. We incorporate its global position within the world coordinate system $\mathbf{p}_{root}$ along with its rotational orientation relative to this global frame $\mathbf{R}_{\text{root}}$. And we employ joint pose of the quadruped robot $\mathbf{q}$ rather than utilizing global positions and velocities of all physical joint points. This approach provides stronger alignment with fundamental robotic control requirements, where root state awareness and joint-space commands constitute essential control inputs. The state $s$ for one robot is shown below:
\begin{equation}
\begin{split}
s=\{\mathbf{p}_{root},  \mathbf{R}_{\text{root}},  \mathbf{q}\}
\end{split}
\end{equation}

\subsubsection{Motion Generation Network}

The motion generation module of our approach is inspired by the foundational architecture proposed by \cite{song2022rsmt}. While retaining the core network structure of the original model, which demonstrated efficacy in generating humanoid motion sequences, we significantly adapt the methodology to address the distinct challenges of quadruped robotic locomotion. Our motion genetator composed three net: PAE network, CMoEs network, and sampler network.



\textbf{PAE Network} First, a periodic autoencoder (PAE) \cite{starke2022deepphase} is trained to generate a multi-dimensional phase manifold. The PAE network decompose motion into periodic components, obtaining an encoding phase manifold, which is a crucial element for training our motion generator.

For application to quadruped robots, our modified PAE architecture replaces the original PAE's state-estimated 3D skeletal velocity inputs with directly obtainable, high-accuracy joint angular velocities. This eliminates the need for costly and error-prone external state estimation pipelines. The dimensionality of the latent space $I$ is set to 5, compared to the 10 dimensions typically required for human motions.


\textbf{CMoEs Decoder} CMoEs Decoder is used to learn different local style motions, which enables the network to generate specified style motions for the motion generator. The network structure is shown in Fig. \ref{Figure: outline}. 

The proposed CMoEs Decoder Network employs an autoencoder structure to train the CMoEs decoder. The network used the current frame $s^t$ and the next frame $s^{t+1}$ to generate the latent $z$. To better emphasize the motion style representation during training, the input Dataset is processed by a trained PAE network to generate a Phase Manifold $\mathcal{P}$. The phase value $p^{t+1}$ extracted from $\mathcal{P}$ at timestep $t+1$ serves as input to a gating network. After that the gating network's output is concatenated with the current state $s_t$, target velocity $v^{t+1}$, and latent variable $z$. This combined vector is fed into an expert pool network to predict the $\Delta{s^{t+1}}$.

For the loss of the CMoEs, firstly, we calculate the Kullback-Leibler (KL) divergence as a loss to constrain the distribution of the latent vector:
\begin{equation}
\mathcal{L}_{\text{KL}} = \frac{1}{2} \sum_{i=1}^{d} \left( \sigma_i^2 + \mu_i^2 - 1 - \ln(\sigma_i^2) \right)
\end{equation}
Here, $d$ is the dimension of the latent space, $\mu_i$ is the posterior mean of the i-th dimension and $\sigma_i$ is the posterior standard deviation of the i-th dimension. And we have a foot loss which is used to measure the foot skating of the motion:
\begin{equation}
\mathcal{L}_{foot}=v_{foot}, \text{when } h_{foot}<\delta
\end{equation}
As for the position loss, we have pos loss ($\mathcal{L}_{\text{pos}}$), joint rotation loss ($\mathcal{L}_{\text{rot}}$), orientation loss ($\mathcal{L}_{\text{ori}}$) and root position loss ($\mathcal{L}_{\text{root}}$), which is shown in Eq. \ref{eq:loss}, where $\mathbf{P}$ is the 3D global joint point calculated by forward kinematic from $s$, $\boldsymbol{\theta}$ is the rotations of robot joints, $\mathbf{R}$ is the root rotation under the global coordinate (6D representation), $\mathbf{r}$ is the root position, and $N$ is the Batch size. This loss design enables dual-constraint optimization: enforcing global coordinate requirements while maintaining robot-specific control requirements. Crucially, it incorporates penalties $\mathcal{L}_{\text{joint limit}}$ for joint limit violations when predicted motions exceed hardware capabilities. Finally we have:
\begin{equation}
\begin{split}
\mathcal{L}_{\text{pos}} = \frac{1}{N \times T} \sum_{i=1}^{N} \sum_{t=1}^{T} \| \mathbf{P}^{(i,t)}_{\text{gt}} - \mathbf{P}^{(i,t)}_{\text{pred}} \|_2^2 \\
\mathcal{L}_{\text{rot}} = \frac{1}{N \times T} \sum_{i=1}^{N} \sum_{t=1}^{T} \| \boldsymbol{\theta}^{(i,t)}_{\text{gt}} - \boldsymbol{\theta}^{(i,t)}_{\text{pred}} \|_2^2 \\
\mathcal{L}_{\text{ori}} = \frac{1}{T} \sum_{t=1}^{T} \| \mathbf{R}^{(t)}_{\text{gt}} - \mathbf{R}^{(t)}_{\text{pred}} \|_2^2 \\
\mathcal{L}_{\text{root pos}} = \frac{1}{T} \sum_{t=1}^{T} \| \mathbf{r}^{(t)}_{\text{gt}} - \mathbf{r}^{(t)}_{\text{pred}} \|_2^2
\label{eq:loss}
\end{split}
\end{equation}
 
\begin{equation}
\begin{split}
\mathcal{L}=\mathcal{L}_{\text{foot}}+\mathcal{L}_{\text{pos}}+\mathcal{L}_{\text{KL}}+\mathcal{L}_{\text{rot}}+\mathcal{L}_{\text{ori}}+\mathcal{L}_{\text{root pos}}+\mathcal{L}_{\text{joint limit}}
\end{split}
\end{equation}

\textbf{Sampler Network} The CMoEs network can predict $\hat{s}^{t+1}$ in the desired style with state $s^t$ and velocity $v^{t+1}$. However, while proficient in predicting the intended style, the output is not able to follow the targeted destination. In order to reach our purpose, a sampler network is needed. The network is shown in Fig. \ref{Figure: outline}.

The proposed motion generation framework takes the current state $s^t$, the target state $s^T$ and the current phase vector $p^t$ of state $s^t$ as input. For the input of the encoder, $s^t$ represents the current motion information, $s^T$ contains the target motion information and $s^T-s^t$ includes the message of the distance between $s^t$ and $s^T$. The encoder takes messages from the input, and transfer them into a latent embedding $\mathbf{z}_{\text{in}}$ as the input of a long short-term memory (LSTM). Then, the LSTM predictor takes the latent embedding with the message of $s^t$ and $s^T$ and predict the latent message of $s^{t+1}$. Finally, the decoder takes the predicted latent message and the phase vector, which represent the style to predict the style of the next frame $p^{t+1}$, the next latent $z^{t+1}$ and the velocity of the next frame $v^{t+1}$. The output of the sampler network will be the input of the CMoEs decoder to generate the next state $s^{t+1}$.

When deploying the algorithm on robotic systems, acquiring the phase vector for the initial frame becomes a critical challenge. In action generation algorithms applied to test datasets, phase vectors for each frame can be directly computed through dataset preprocessing. However, in actual implementation scenarios—where only the starting frame and target frame data are initially available—calculating phase vectors requires a sequence of consecutive frames. We address this by employing a deep learning network to predict the initial phase vector. Specifically, we construct a frame sequence by repeatedly replicating the starting frame to provide necessary temporal context for the convolutional layers. This frame sequence is then processed by the neural network architecture described in Table \ref{tab:Init Phase Prediction Network Architecture} and the first frame predict part in Fig. \ref{Figure: outline} to predict the initial phase vector.

\begin{table}[h]
\centering
\caption{Init Phase Prediction Network Architecture}
\label{tab:Init Phase Prediction Network Architecture}
\begin{tabular}{c|c|c}
\hline
\textbf{Module} & \textbf{Layers} & \textbf{Parameters} \\
\hline
Feature Extraction 1 & 
\begin{tabular}{@{}c@{}}
Conv1d \\ 
BN \\ 
ReLU \\ 
Dropout
\end{tabular} &
\begin{tabular}{@{}c@{}}
Kernel=5 \\ 
Padding=2 \\ 
Channels=1024 \\ 
Drop=0.4
\end{tabular} \\
\hline
Feature Extraction 2 & 
\begin{tabular}{@{}c@{}}
Conv1d \\ 
BN \\ 
ReLU \\ 
Dropout
\end{tabular} &
\begin{tabular}{@{}c@{}}
Kernel=3 \\ 
Padding=1 \\ 
Channels=1024 \\ 
Drop=0.4
\end{tabular} \\
\hline
Output Conv & 
\begin{tabular}{@{}c@{}}
Conv1d \\ 
BN \\ 
ReLU
\end{tabular} &
\begin{tabular}{@{}c@{}}
Kernel=3 \\ 
Padding=1 \\ 
Channels=1024
\end{tabular} \\
\hline
Phase Mapping & 
\begin{tabular}{@{}c@{}}
Linear \\ 
Output
\end{tabular} &
\begin{tabular}{@{}c@{}}
In=1024 \\ 
Out=$\text{PhaseDim}  \times 3$
\end{tabular} \\
\hline
\end{tabular}
\end{table}

\subsection{Adversarial Imitation Learning based robot Control}

This method learns agile and controllable legged locomotion policies by integrating Adversarial Imitation Learning (AIL) with task objectives. A data-driven \textit{motion prior}, acquired through adversarial training, regularizes policy behavior to produce natural, stable, and energy-efficient motions suitable for sim-to-real transfer.

\textbf{Problem Formulation \& Markov Decision Process (MDP)} Legged locomotion learning is modeled as an MDP: $(\mathcal{S},\mathcal{A}, f, r_{t}, p_{0},\gamma)$. $\mathcal{S}$ denotes the state space, $\mathcal{A}$ the action space, $f(s, a)$ the system dynamics, $r_{t}(s, a, s^{\prime})$ the reward function, $p_{0}$ the initial state distribution, and $\gamma$ the discount factor. The Reinforcement Learning (RL) objective is to find optimal parameters $\theta$ for policy $\pi_{\theta}:\mathcal{S}\mapsto\mathcal{A}$ that maximize the expected discounted return:
\begin{equation}
\begin{split}
J(\theta)=E_{\pi_{\theta}}\left[\sum_{t=0}^{T-1}\gamma^{t} r_{t}\right]
\end{split}
\end{equation}

\textbf{Observations} The policy observation vector \(\mathbf{o}_{t}\) integrates multimodal state information:

\begin{equation}
\begin{split}
\mathbf{o}_{t} = 
\begin{bmatrix}
    \mathbf{q},\ \dot{\mathbf{q}},
    \mathbf{G_p},
    \boldsymbol{\omega},
    \mathbf{C}, 
    \mathbf{a}_{t-1}
\end{bmatrix}
\end{split}
\end{equation}

This includes: 12-dimensional joint positions \(\mathbf{q} \in \mathbb{R}^{12}\) and velocities \(\dot{\mathbf{q}} \in \mathbb{R}^{12}\), projected gravity \(\mathbf{G_p} \in \mathbb{R}^{3}\), base angular velocity \(\boldsymbol{\omega} \in \mathbb{R}^{3}\), command \(\mathbf{C} \in \mathbb{R}^{6}\), including the base velocity command and angular velocity command, and previous action \(\mathbf{a}_{t-1} \in \mathbb{R}^{12}\).

During simulation training, privileged information \(\mathbf{v}_{\text{base}}\) (base linear velocity) is incorporated to form an augmented observation vector, which is excluded during real-world implementation.

\textbf{Action Generation and Motion Initialization Mechanism} The policy outputs joint position deltas, and the executed joint pose is obtained by superimposing the policy-generated delta onto a predefined default configuration:

\begin{equation}
\Delta\mathbf{q} = \pi_{\theta}(\mathbf{o}_{t}) \\
\mathbf{q}_{\text{exec}} = \mathbf{q}_{\text{default}} + \Delta\mathbf{q}
\end{equation}

\textbf{Reward Function} Our reward function comprises three components: task reward $r_{t}^{T}$, regularization reward $r_{t}^{R}$, and style reward $r_{t}^{S}$, and the complete reward formulation is summarized in Table \ref{tab: Reward terms for training policy}.

\begin{table}[h]
\centering
\caption{Reward terms for training policy}
\label{tab: Reward terms for training policy}
\renewcommand{\arraystretch}{1.3}
\begin{tabular}{c|c|c|c}
\hline
\textbf{Term} & \textbf{Weight} & \textbf{Equation} & \textbf{Scale} \\
\hline
Task $r^{T}_{t}$ &
\begin{tabular}{@{}c@{}}
0.2
\end{tabular} &
\begin{tabular}{@{}c@{}}
$\exp\{-4 (\boldsymbol{v}_{xy}^{cmd}-\boldsymbol{v}_{xy})^2\}$ \\ 
$\exp\{-4(\boldsymbol{\omega}^{cmd}_{z}-\boldsymbol{\omega}_{z})^2\}$ 
\end{tabular} &
\begin{tabular}{@{}c@{}}
1.0 \\ 
0.5 \\ 
\end{tabular} \\
\hline
Style $r^{S}_{t}$ & 
\begin{tabular}{@{}c@{}}
0.8
\end{tabular} &
\begin{tabular}{@{}c@{}}
$r_{t}^{s}(s_{t},s_{t+1})$
\end{tabular} &
\begin{tabular}{@{}c@{}}
1.0
\end{tabular} \\
\hline
\begin{tabular}[c]{@{}c@{}} 
Regularization \\ 
${r}^{R}_{t}$
\end{tabular} &
\begin{tabular}{@{}c@{}}
0.5
\end{tabular} &
\begin{tabular}{@{}c@{}}
$||\boldsymbol{\tau}||^2$ \\
$||\boldsymbol{a}_{t}-\boldsymbol{a}_{t-1}||^2$ \\
$||\boldsymbol{\ddot{q}}||^2$ \\
$||\max \left[0,\left|\boldsymbol{\tau}\right|- \boldsymbol{\tau^{lim}}\right]||^2$\\
$\sum_{f}^{4}(t_{air,f}-0.5)$\\
\end{tabular} &
\begin{tabular}{@{}c@{}}
$-1 \times 10^{-5}$\\ 
-0.01\\
$-2.5 \times 10^{-7}$\\ 
$-5 \times 10^{-5}$ \\
$ 1.0$
\end{tabular} \\
\hline
\end{tabular}
\end{table}

\section{Result}


\subsection{Motion Generation Results Analysis}

\begin{figure}[htbp] 
    \subfloat[Gallop]{ 
    \begin{minipage}[b]{0.5\textwidth}
        \centering  
        \includegraphics[width=0.75\linewidth]{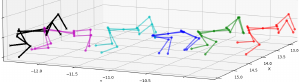} 
    \end{minipage}} 
    \\
    \subfloat[Tripod]{ 
    \begin{minipage}[b]{0.5\textwidth}
        \centering  
        \includegraphics[width=0.75\linewidth]{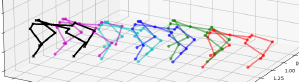} 
    \end{minipage}} 
    \\
    \subfloat[Trotting]{ 
    \begin{minipage}[b]{0.5\textwidth}
        \centering  
        \includegraphics[width=0.75\linewidth]{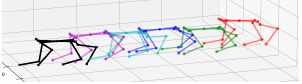} 
    \end{minipage}} 
    \\
       \subfloat[Pacing]{
    \begin{minipage}[b]{0.5\textwidth}
        \centering  
        \includegraphics[width=0.75\linewidth]{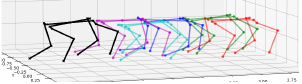} 
    \end{minipage}} 
    \caption{Experimental Results of Motion Generation} 
    \label{Figure: Result Sampler Network} 
\end{figure}

This proposed network synthesizes intermediate frames between specified start and end poses while preserving quadruped gait style characteristics. Fig. \ref{Figure: Result Sampler Network} comprehensively demonstrates this process across four distinct locomotion patterns: gallop, tripod, trotting, and pacing.

The results show essential gait features: gallop sequences exhibit characteristic aerial posture, tripod motions maintain consistent triangular support configurations across frames, trotting patterns preserve diagonal limb coordination, and pacing demonstrates same-side leg synchronization. The generated motion sequences demonstrate robust performance across all four gait types.

For comparison, we compare our method with RSMT \cite{song2022rsmt}. Our comparative analysis focuses on motion tracking accuracy, which is the most critical metric for evaluating action generation systems. To ensure equitable benchmarking, we retrained the RSMT network using identical quadruped robot skeletal configurations and training protocols. The result is shown in Table. \ref{Table: comparison table of rsmt}: the terminal frame position deviation and full trajectory consistency.

\begin{table}[htbp] 
    \centering
    \caption{Comparison on the L2 norm of global position of the last predicted frame and the overall predicted frame on test set}
    \label{Table: comparison table of rsmt}
    \begin{tabular}{c|c|c|c|c}
    \hline
        \multicolumn{5}{c}{L2 norm of global position of the last frame}\\
        \hline
        Quadruped Gait             & Gallop &Tripod &Trotting &Pacing              \\
        \hline
        RSMT                  &   0.727       &   0.471      &  0.444     &  0.437    \\
        Our Method              & \textbf{0.214}  & 0.465   &  \textbf{0.344} & \textbf{0.329} \\
        \hline 
        \multicolumn{5}{c}{} \\
        \hline
        \multicolumn{5}{c}{L2 norm of global position of a motion clips in test set} \\
        \hline
        Quadruped Gait             & Gallop &Tripod &Trotting &Pacing              \\
        \hline
        RSMT                  &   0.501       &  0.450       &  0.497     &  0.277    \\
        Our Method              & \textbf{0.186} &  \textbf{0.058} &  \textbf{0.135} &\textbf{0.054} \\

        \hline
    \end{tabular}
\end{table}

Results demonstrate our method's consistent superiority across nearly all evaluations. While matching baseline terminal accuracy during trotting, we outperform in all other gaits. This establishes advanced capability in maintaining precise long-duration trajectory tracking for robots grappling with error accumulation and dynamic balancing. Our approach delivers critical deployment value through reliable continuous motion generation.

Our approach preserves canonical quadrupedal configurations, enabling direct robotic imitation learning integration without motion retargeting or kinematic pre-processing. This structural fidelity eliminates computational overhead from coordinate transformations and validity screening, significantly accelerating both training and deployment.

\subsection{Imitation Learning Results Analysis}

\begin{figure}[htbp] 
    \centering
    \subfloat[Gallop]{ 
    \begin{minipage}[b]{0.5\textwidth}
        \centering  
        \includegraphics[width=0.9\linewidth]{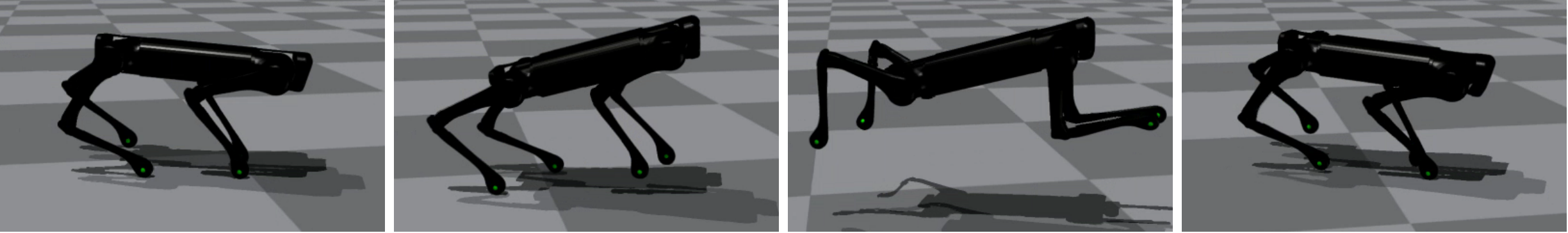} 
    \end{minipage}} 
    \\
    \subfloat[Tripod]{ 
    \begin{minipage}[b]{0.5\textwidth}
        \centering  
        \includegraphics[width=0.9\linewidth]{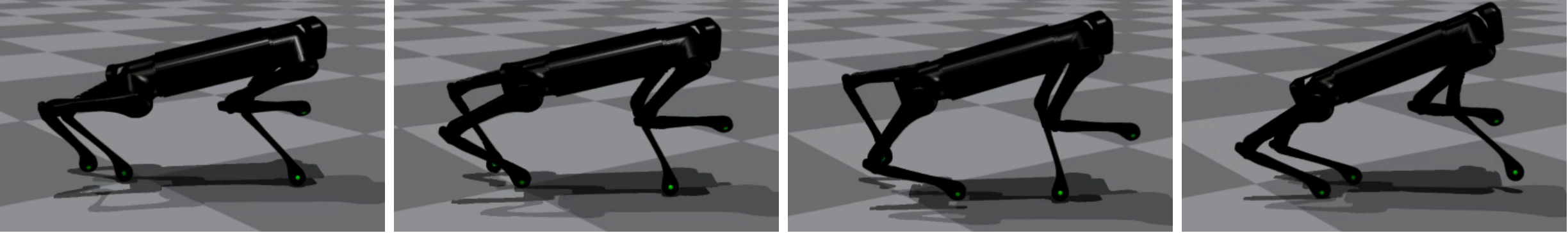} 
    \end{minipage}} 
    \\
    \subfloat[Trotting]{ 
    \begin{minipage}[b]{0.5\textwidth}
        \centering  
        \includegraphics[width=0.9\linewidth]{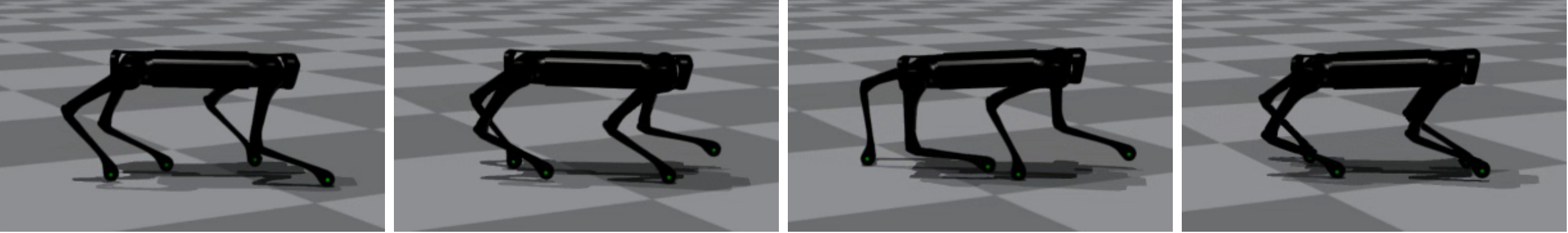} 
    \end{minipage}} 
    \\
       \subfloat[Pacing]{ 
    \begin{minipage}[b]{0.5\textwidth}
        \centering  
        \includegraphics[width=0.9\linewidth]{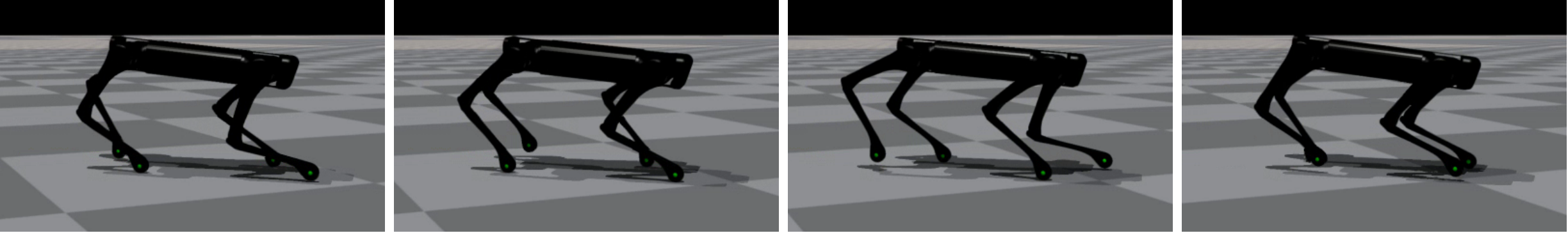} 
    \end{minipage}} 
    \caption{Multi-Motion Adversarial Imitation Results in Isaac Gym.} 
    \label{Figure: Result of AMP in isaacgym} 
\end{figure}

Fig. \ref{Figure: Result of AMP in isaacgym} temporally demonstrates the imitation effects across a full locomotion cycle of four distinct gaits.

The visualization particularly highlights two dynamically challenging gaits: gallop requiring aerial phases with precise landing coordination, and tripod demanding continuous stability maintenance while keeping the left forefoot elevated. Successful execution of these complex maneuvers evidences the inherent stability properties embedded within the generated motion data. This motion stability proves essential for effective integration with AMP training, as the kinematically consistent and dynamically balanced references provide ideal learning targets for imitation learning pipelines.

The trotting and pacing gaits represent more tractable locomotion patterns that do not require extensive dynamic maneuvers, making them comparatively straightforward to achieve through training. Our simulations successfully demonstrate the generation and execution of both gaits with AMP, completing the comprehensive showcase of quadrupedal locomotion capabilities.

\begin{figure}[htbp] 
    \centering
    \subfloat[Gallop]{ 
    \begin{minipage}[b]{0.5\textwidth}
        \centering  
        \includegraphics[width=0.9\linewidth]{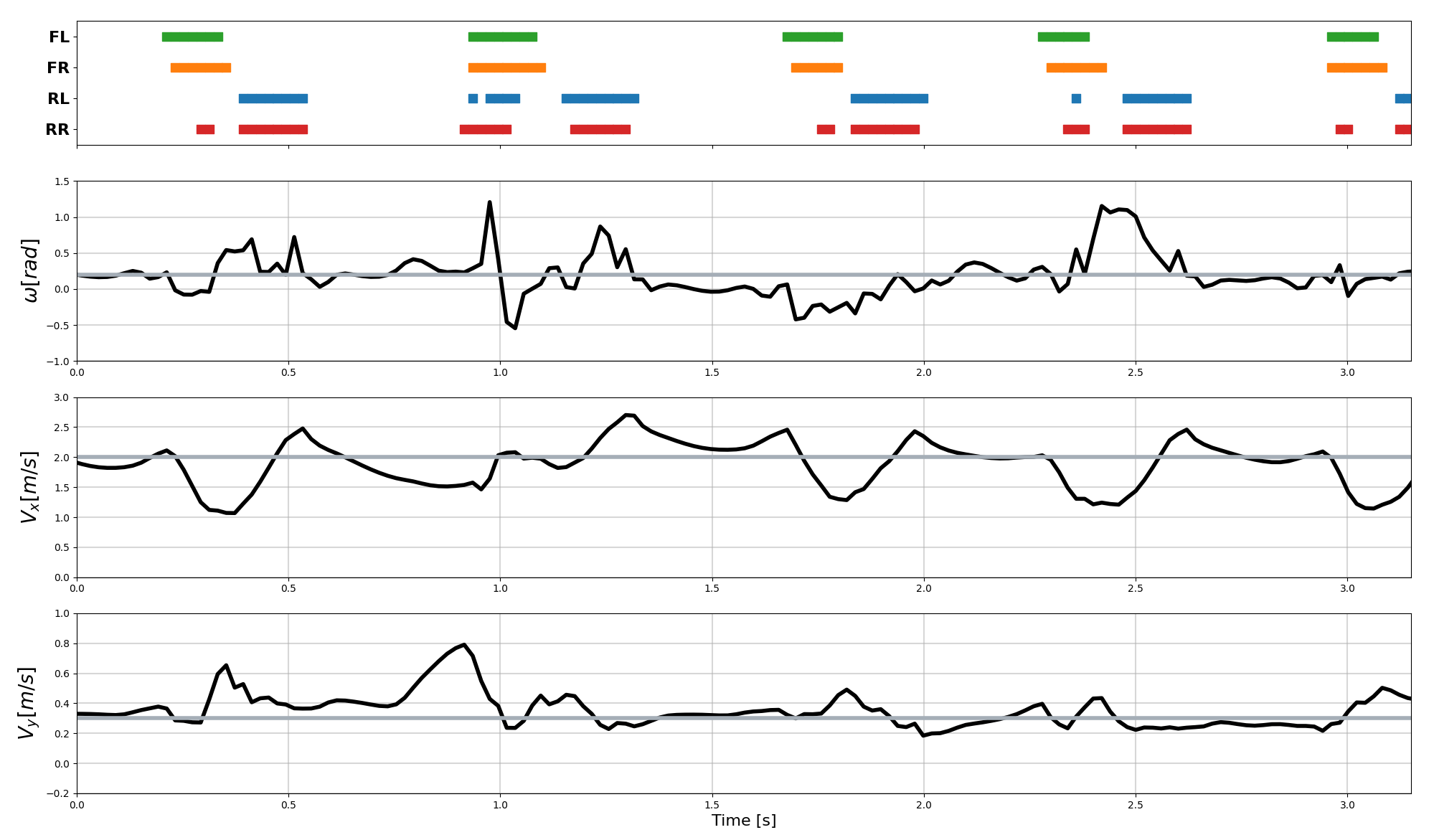} 
    \end{minipage}} 
    \\
    \subfloat[Tripod]{ 
    \begin{minipage}[b]{0.5\textwidth}
        \centering  
        \includegraphics[width=0.9\linewidth]{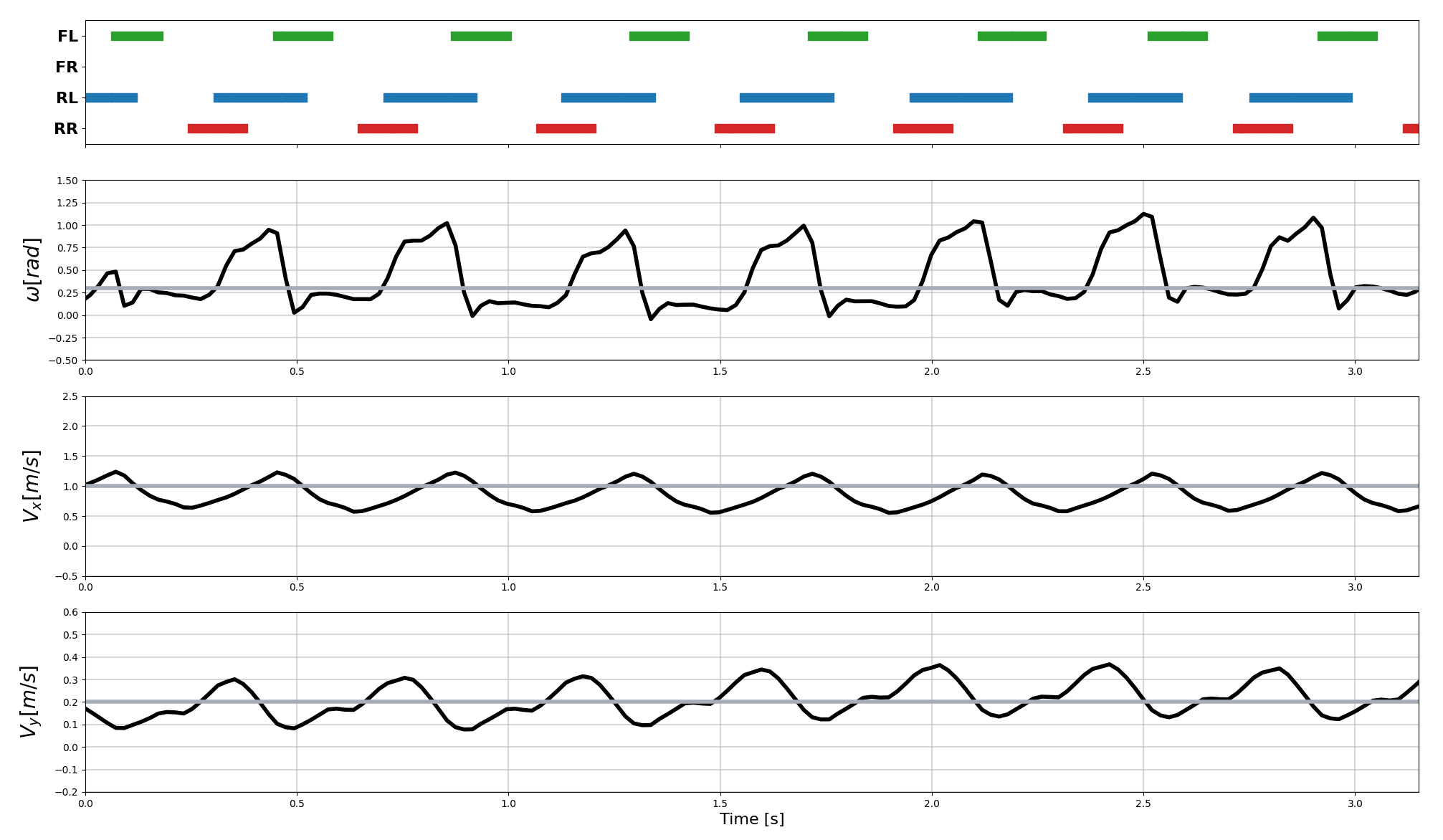} 
    \end{minipage}} 
    \\
    \subfloat[Trotting]{ 
    \begin{minipage}[b]{0.5\textwidth}
        \centering  
        \includegraphics[width=0.9\linewidth]{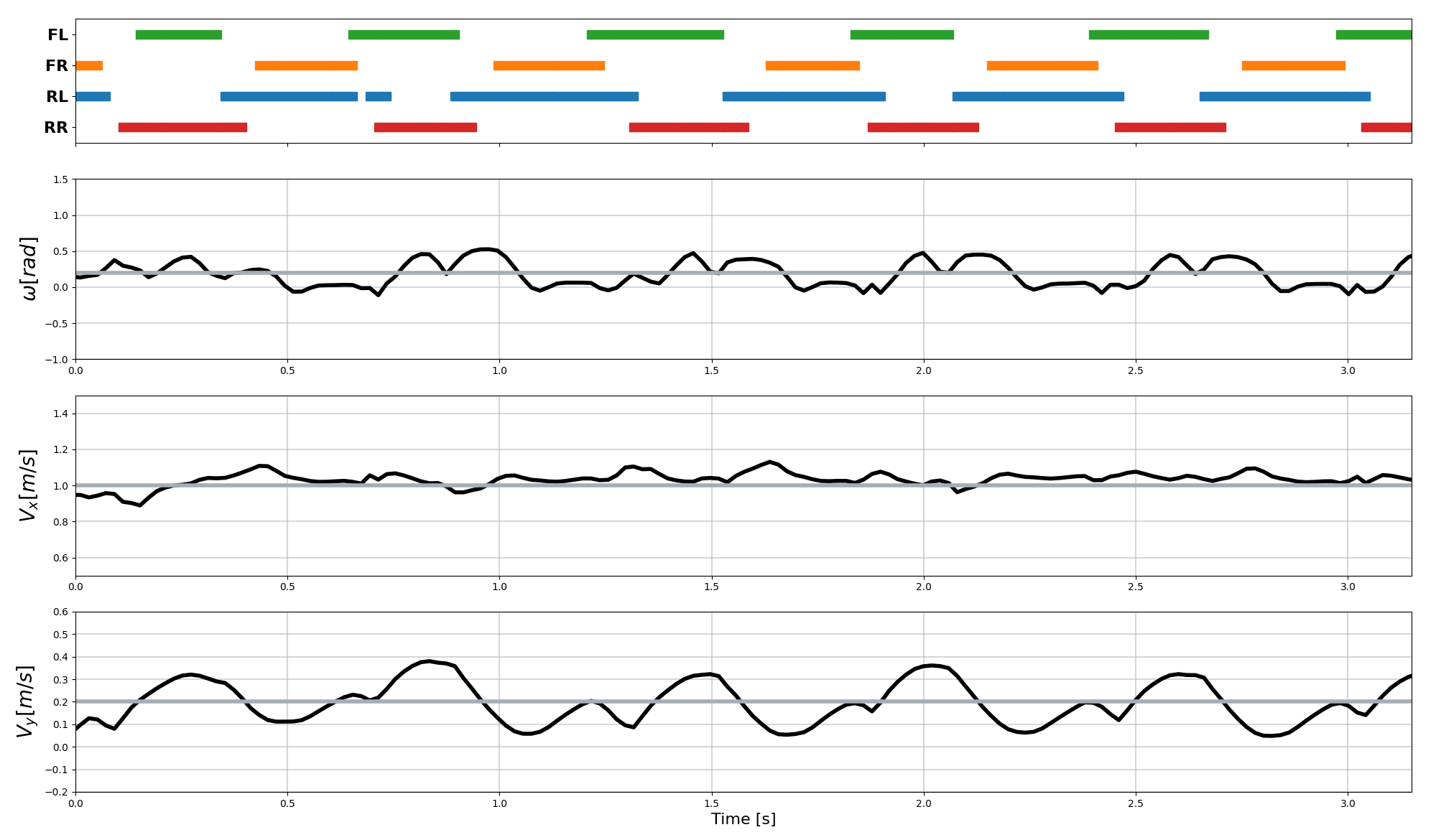} 
    \end{minipage}} 
    \\
       \subfloat[Pacing]{ 
    \begin{minipage}[b]{0.5\textwidth}
        \centering  
        \includegraphics[width=0.9\linewidth]{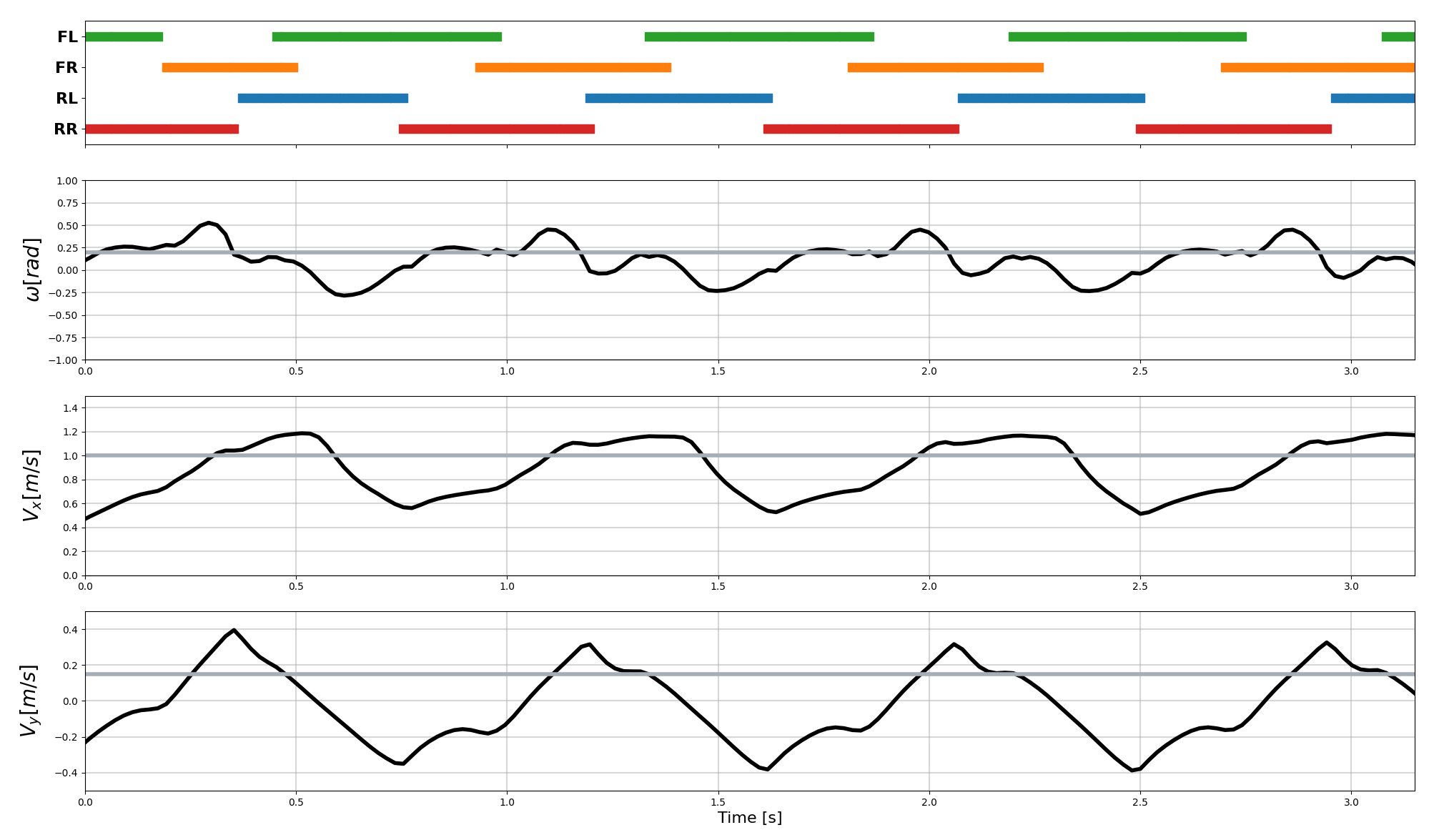} 
    \end{minipage}} 
    \caption{Command Tracking Performance: first line: Gait phase analysis. 
        Second line: Angular velocity ($\omega$) tracking performance with reference commands in gray. 
        Third line: Translational velocity tracking along robot X-axis ($v_x$).  
        Bottom: Translational velocity tracking along robot Y-axis ($v_y$).} 
    \label{Figure: Command Result of AMP in isaacgym} 
\end{figure}

To evaluate the robot's velocity command tracking performance, we analyzed its command-following capabilities in a simulation environment of four representative locomotion gaits (gallop, tripod, pacing, and trotting) under specified reference velocity commands. The corresponding command tracking trajectories are presented in Figs. \ref{Figure: Command Result of AMP in isaacgym}.

\begin{figure}[htbp] 
    \centering
    \subfloat[Gallop]{ 
    \begin{minipage}[b]{0.5\textwidth}
        \centering  
        \includegraphics[width=0.85\linewidth]{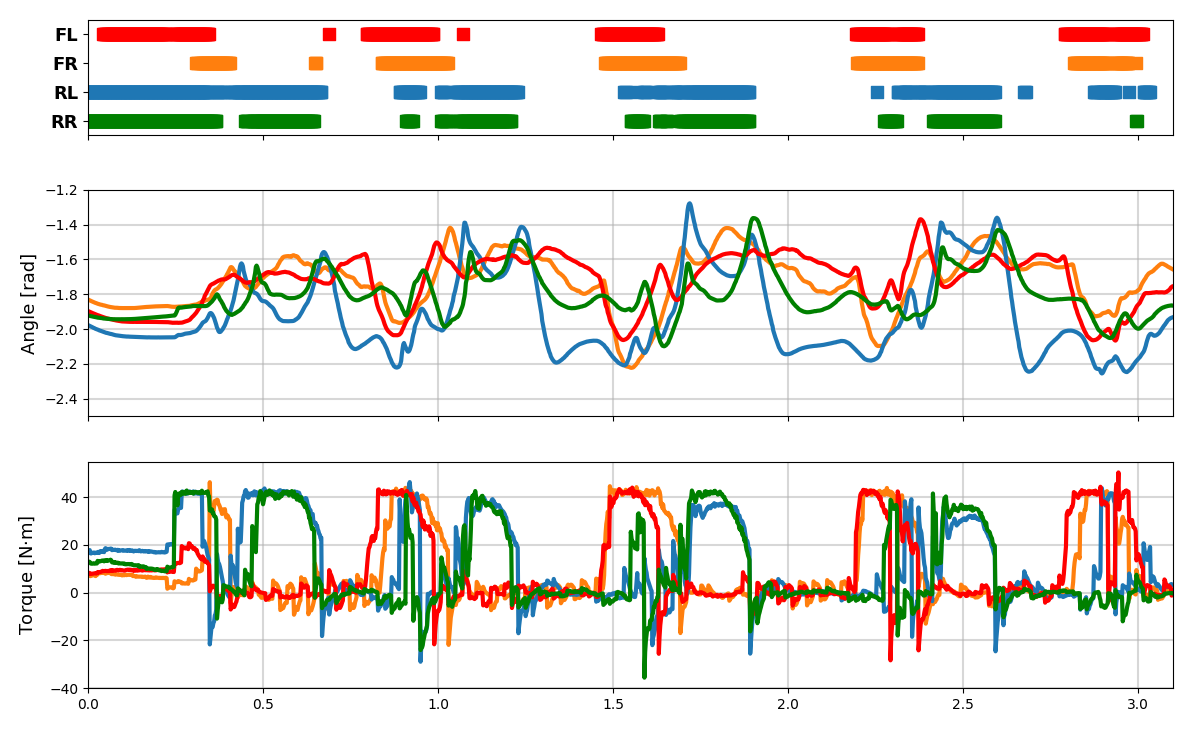} 
    \end{minipage}} 
    \\
       \subfloat[Tripod]{ 
    \begin{minipage}[b]{0.5\textwidth}
        \centering  
        \includegraphics[width=0.85\linewidth]{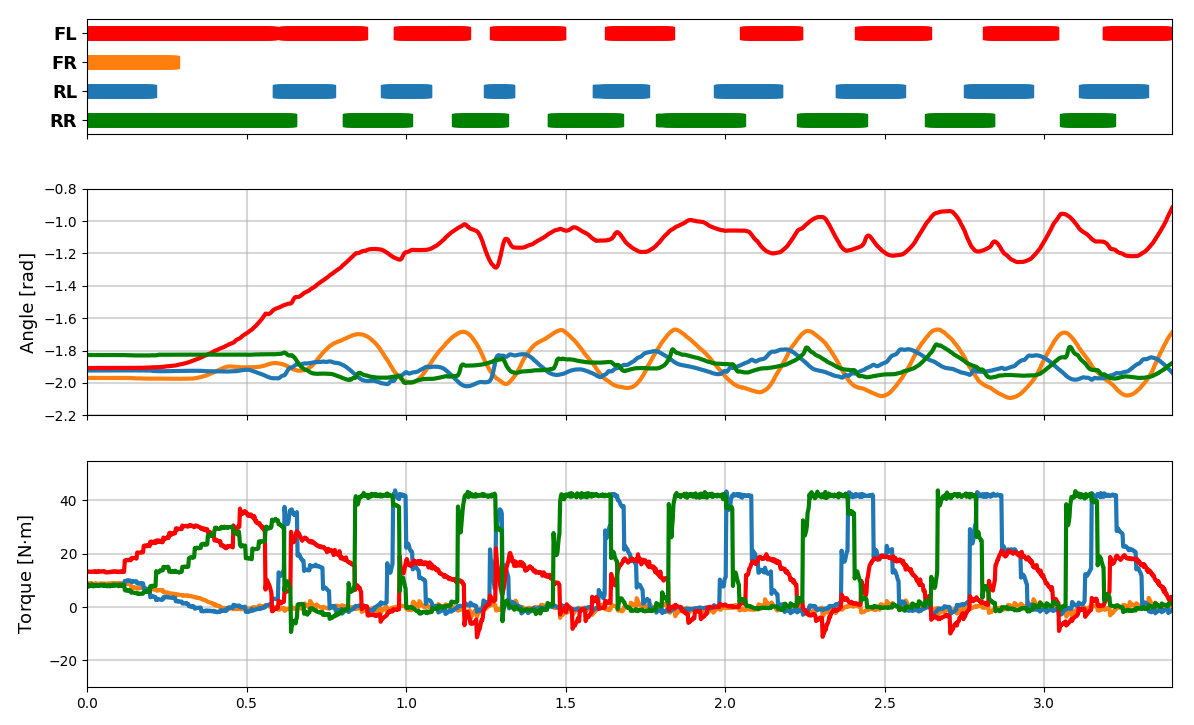} 
    \end{minipage}} 
    \caption{Motion data collected during real-world experiment on Unitree AlienGo. First Line: Gallop gait Analysis, the lines represent the feet contact status of FL(front left), FR(front right), RL(rear left), RR(rear right) feet. Second Line: Calf joints angle over time. Third Line: Calf joints torque over time.} 
    \label{Figure: Data of Deployment} 
\end{figure}

In real-world robot testing, we focused on validating gallop and tripod gaits, which represent challenging locomotion. The results is shown in Fig. \ref{Figure: Result of AMP in real world}, and the corresponding trajectories data are presented in Figs. \ref{Figure: Data of Deployment}. Tripod gait requires keeping one leg always lifted while the other three legs form a stable triangle support, demanding excellent balance control. Gallop involves jumping motions with flight phases, testing the robot's ability to handle strong impacts and deliver high power.

Experimental results matched what we saw in simulations: during gallop, front legs touched down first to absorb impact and store energy, followed by the back legs pushing powerfully. Joints straightened during jumps and bent when landing. We measured a stronger foot pushing force during takeoff and clear flight phases between steps. Tripod execution maintained constant tripedal support through rapid leg alternation, necessitating faster stepping cycles and continuous balance compensation compared to gallop.

Successfully executing both gaits proves our motion generation method works in real-world conditions. This shows our system can create both big explosive movements like gallop and precise balancing motions like tripod, while keeping the robot stable enough for reliable real-world deployment after simulation training.

\section{CONCLUSIONS}


In summary, we propose an in-between motion generation based multi-style quadruped robot locomotion framework, which is capable of generating motions at arbitrary velocities using sparse motion capture data, while achieving accurate velocity tracking performance through imitation learning based deployment on real-world robots. Experimental results demonstrate that this approach significantly enhances quadrupedal locomotion performance across velocity tracking and dynamic stability metrics, which successfully solves the lack of data problem in imitation learning.










\balance
\bibliographystyle{IEEEtran}
\bibliography{ref}

\end{document}